\documentclass[10pt, a4paper]{article}

\usepackage{lrec-coling2024} 

\usepackage{natbib}
\usepackage{multibib}
\makeatletter
\def\@mb@citenamelist{cite,citep,citet,citealp,citealt,citepalias,citetalias}
\makeatother
\newcites{languageresource}{~}

\usepackage{graphicx}
\usepackage{soul}
\usepackage{dsfont}
\usepackage{bbm}
\usepackage{amsmath}
\usepackage{booktabs}
\usepackage{amsfonts}

\usepackage{enumitem}
\usepackage{caption}
\usepackage{multicol}
\usepackage{multirow}

\usepackage{float}
\usepackage{efbox}
\usepackage{subfigure}
\usepackage{hyperref}
\usepackage{cleveref}
\usepackage{colortbl}
\usepackage{url}
\usepackage{amssymb}
\usepackage{pifont}
\usepackage[ruled, lined, linesnumbered, commentsnumbered, longend]{algorithm2e}

\usepackage{titlesec}
\titleformat{\section}{\normalfont\large\bfseries\center}{\thesection.}{1em}{}
\titleformat{\subsubsection}{\normalfont\normalsize\bfseries\raggedright}{\thesubsubsection.}{1em}{}
\renewcommand\thesection{\arabic{section}}
\renewcommand\thesubsection{\thesection.\arabic{subsection}}
\renewcommand\thesubsubsection{\thesubsection.\arabic{subsubsection}}
\newcommand{\ourmethod}{RAP}

\definecolor{lightgray}{rgb}{0.9,0.9,0.9}

\usepackage{hyperref}
 \definecolor{darkblue}{rgb}{0, 0, 0.5}
  \hypersetup{colorlinks=true, citecolor=darkblue, linkcolor=darkblue, urlcolor=darkblue}

\usepackage{xstring}

\usepackage{color}

\title{New Intent Discovery with Attracting and Dispersing Prototype}

\name{Shun Zhang$^{1, 2}$, Jian Yang$^{1}$$^{*}$\thanks{$^{*}$Corresponding author.}, Jiaqi Bai$^{1, 2}$, Chaoran Yan$^{1}$, 
\\{\bf \large Tongliang Li$^{3}$, Zhao Yan$^{4}$,  Zhoujun Li$^{1, 2}$}
}

\address{
$^1$ State Key Lab of Software Development Environment, Beihang University, Beijing, China\\
$^2$ School of Cyber Science and Technology, Beihang University, Beijing, China \\
$^3$ Beijing Information Science and Technology University, $^4$ Tencent Youtu Lab\\
\texttt{\{shunzhang, jiaya,bjq,ycr2345,lizj\}@buaa.edu.cn}\\ 
\texttt{\{tonyliangli\}@bistu.edu.cn};\texttt{\{zhaoyan\}@tencent.com}
}

\abstract{
New Intent Discovery (NID) aims to recognize known and infer new intent categories with the help of limited labeled and large-scale unlabeled data. The task is addressed as a feature-clustering problem and recent studies augment instance representation. However, existing methods fail to capture cluster-friendly representations, since they show less capability to effectively control and coordinate within-cluster and between-cluster distances.
Tailored to the NID problem, we propose a \textbf{R}obust and \textbf{A}daptive \textbf{P}rototypical learning (\ourmethod{}) framework for globally distinct decision boundaries for both known and new intent categories. Specifically, a robust prototypical attracting learning (RPAL) method is designed to compel instances to gravitate toward their corresponding prototype, achieving greater within-cluster compactness.
To attain larger between-cluster separation, another adaptive prototypical dispersing learning (APDL) method is devised to maximize the between-cluster distance from the prototype-to-prototype perspective. Experimental results evaluated on three challenging benchmarks (CLINC, BANKING, and StackOverflow) of our method with better cluster-friendly representation demonstrate that \ourmethod{} brings in substantial improvements over the current state-of-the-art methods (even large language model) by a large margin (average $+5.5\%$ improvement).
 \\ \newline \Keywords{New Intent Discovery, Robust Prototypical Attracting, Adaptive Prototypical Dispersing} 
}

\begin{document}

\maketitleabstract

\section{Introduction}
Due to the success of conventional intent detection in dialogue systems~\cite {Dialog2,Dialog3,usnid,zhang-2022-new-intent-discovery}, the vast majority of learning algorithms under the closed-world scenario with static data distribution only consider pre-defined intents. To handle the new intents outside the existing known intent categories, it is necessary to equip dialogue systems with new intent discovery (NID) abilities \cite{idas,gid,zero_shot_gid,unified_nid_cv,nid_industry_setting}.

Early works~\cite{hakkani2013weak,hakkani2015clustering,shi-etal-2018-auto,padmasundari2018intent} mainly adopt unsupervised clustering with unlabeled data. They always ignore prior knowledge of the available labeled data and fail to generate highly accurate and granular intent groups, leading to inapplicability in the open-world scenario. Recent studies are adept to semi-supervised settings to efficiently utilize the limited labeled data, such as pairwise similarities~\cite{lin2020discovering}, iterative pseudo-labeling~\cite{zhang2021discovering}, probabilistic architecture~\cite{zhou2023latent} and prototypical network~\cite{DPN_AAAI2023}.
Due to the transferability of injecting structural knowledge from known categories into the intent representation, the semi-supervised methods can be extended to real-world scenarios with better NID performance. 

\begin{figure}[t]   
\centering
\includegraphics[width=1.0\columnwidth]{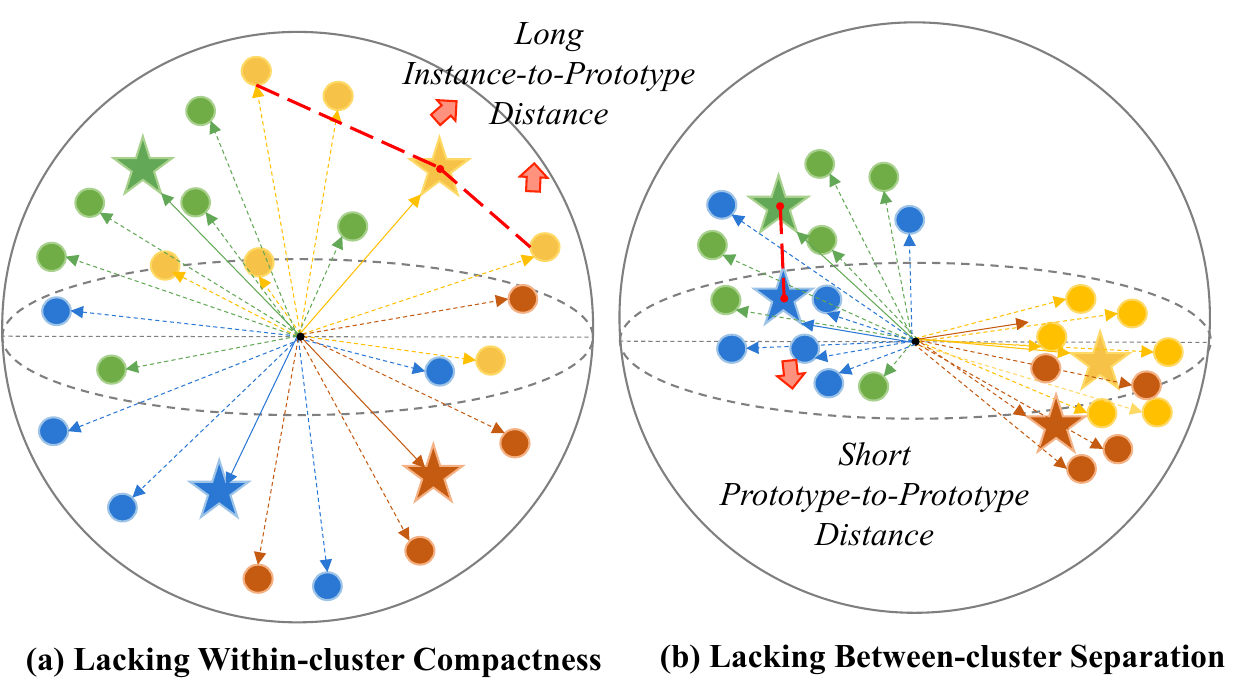}
\caption{Embedding distribution of intent instances and the prototype of each class in a shared sphere semantic space. The circle and star shape denote the instance and the prototype, respectively. The discriminative representations fail to be extracted due to insufficient (a) within-cluster compactness and (b) between-cluster separation.}
\label{fig:fig_1}
\end{figure}

However, existing semi-supervised methods still face two challenges: \textbf{(C1)}: Lacking sufficient within-cluster compactness within the learned intent representations. \textbf{(C2)}: Requiring the explicit modeling of between-cluster dispersion in the representation space. 
In Figure~\ref{fig:fig_1}(a), different intent categories overlap with each other with varying distributions, where the large class-specific variances result in distant instance embeddings from the cluster centers. In Figure~\ref{fig:fig_1}(b), embeddings in different categories tend to gather in the same region, due to the narrow distance among cluster centers. Previous baselines encounter difficulty in capturing clear and accurate cluster boundaries for known and novel categories. Therefore, \emph{Jointly constraining both within- and between-cluster distance to yield cluster-friendly discriminative representations still requires further exploration.}

In this work, we propose a \textbf{R}obust and \textbf{A}daptive \textbf{P}rototypical learning (\ourmethod{}) framework for the joint identification of known intents and discovery of the novel intents. To mitigate the issue of insufficient within-cluster compactness, we introduce a robust prototypical attracting learning (\textbf{RPAL}) method to reduce the overlarge variances from an instance-to-prototype perspective.
Specifically, we first compute class prototypes as normalized mean embeddings and enforce each instance embedding to be closer to its corresponding class prototypes. 
To avoid the negative impact of pseudo-label noise, a novel interpolation training strategy is used to construct virtual training samples for the maintenance of the same linear relationship with its prototypes.
To address the concern of insufficient between-cluster separation, 
we propose an adaptive prototypical dispersing learning (\textbf{APDL}) to explicitly enlarge the between-cluster distance from a prototype-to-prototype perspective.
By minimizing the total prototype-to-prototype similarities, APDL adaptively maximizes the distance between prototypes to form well-separated clusters. A weighted training objective is used to adaptively impose large penalties on nearer prototypes to push them further apart, helping achieve better dispersion among prototypes.
Finally, RPAL and APDL are jointly optimized with multitask learning to guide the model to learn cluster-friendly intent representations for both known and novel intents.

Experimental results on multiple NID benchmarks demonstrate our method brings substantial improvements over previous
state-of-the-art methods by a large margin of $+5.5\%$ points. Our key contributions are summarized as follows:
\begin{itemize}
\item A new prototype-guided learning framework is designed to learn cluster-friendly discriminative representations with stronger within-cluster compactness and larger between-cluster separation.
\item We propose a robust prototypical attracting learning method and an adaptive prototypical dispersion learning method, which solve the problems of insufficient within-cluster compactness and between-cluster separation.
\item Extensive experiments on three benchmark datasets show that our model establishes state-of-the-art performance on the semi-supervised NID task (average $+5.5\%$ improvement), which demonstrates competitive NID performance.
\end{itemize}


\section{Related Work}
\label{Related_Work}
\subsection{New Intent Discovery}
Existing NID methods can be divided into two categories: unsupervised and semi-supervised.
For the former, pioneering works~\cite{hakkani2013weak,hakkani2015clustering} primarily rely on statistical features of the unlabeled data to cluster similar queries for discovering new user intents. Subsequently, some studies~\cite{xie2016unsupervised,yang2017towards,shi-etal-2018-auto} endeavor to leverage deep neural networks to learn robust representations conducive to new intent clustering.
However, these methods lack the capacity to leverage prior knowledge for clustering guidance.
Addressing this limitation, some studies~\cite{basu2004active,hsu2018,hsu2019learning,han2019learning} begin to explore the use of semi-supervised clustering methods to better leverage prior knowledge.
For example, ~\citet{lin2020discovering} combines the pairwise constraints and target distribution to discover new intents while~\citet{zhang2021discovering} introduces an alignment strategy to improve the clustering consistency.
Further, \citet{shen2021semi,NAACL2022,GCD_cvpr2022,zhang-2022-new-intent-discovery} designs contrastive learning strategies in both the pre-training phase and the clustering stage to learn discriminative representations of intents.
Recently, \citet{zhou2023latent} introduces a principled probabilistic framework and \citet{DPN_AAAI2023} proposed a decoupled prototypical network to enhance the performance of the NID. 
However, these methods fail to effectively capture discriminative representations with strong within-cluster compactness and large between-cluster separation.
This difficulty makes it challenging to differentiate between the characteristics of known and novel intents.

\subsection{Prototypical Learning}
Prototypical learning (PL) methods~\cite{PL2017} have become promising approaches due to their simplicity and effectiveness and they have been widely applied in various scenarios, such as unsupervised domain adaptation~\cite{domain_adaptation}, out-of-domain detection~\cite{DASFAA_zhang,ICASSP_zhang}, machine translation~\cite{smart_start,low_resouce,learn_to_select,yang2020improving,chai2024xcot}, and named entity recognition~\cite{NER_prototype,CROP_Yang_2022,mo2023mcl}.
Among them, prototypical contrastive learning~\cite{PCL_ICLR} is proposed to generate compact clusters. 
It employs cluster centroids as prototypes and trains the network by drawing instance embeddings closer to its assigned prototypes. 
Here, we explore the utilization of the interpolation training strategy to enhance cluster compactness while mitigating the effects of label noise, thus rendering it more robust.


\section{Approach}
\begin{figure*}[t]   
\centering
\includegraphics[width=0.80\linewidth]{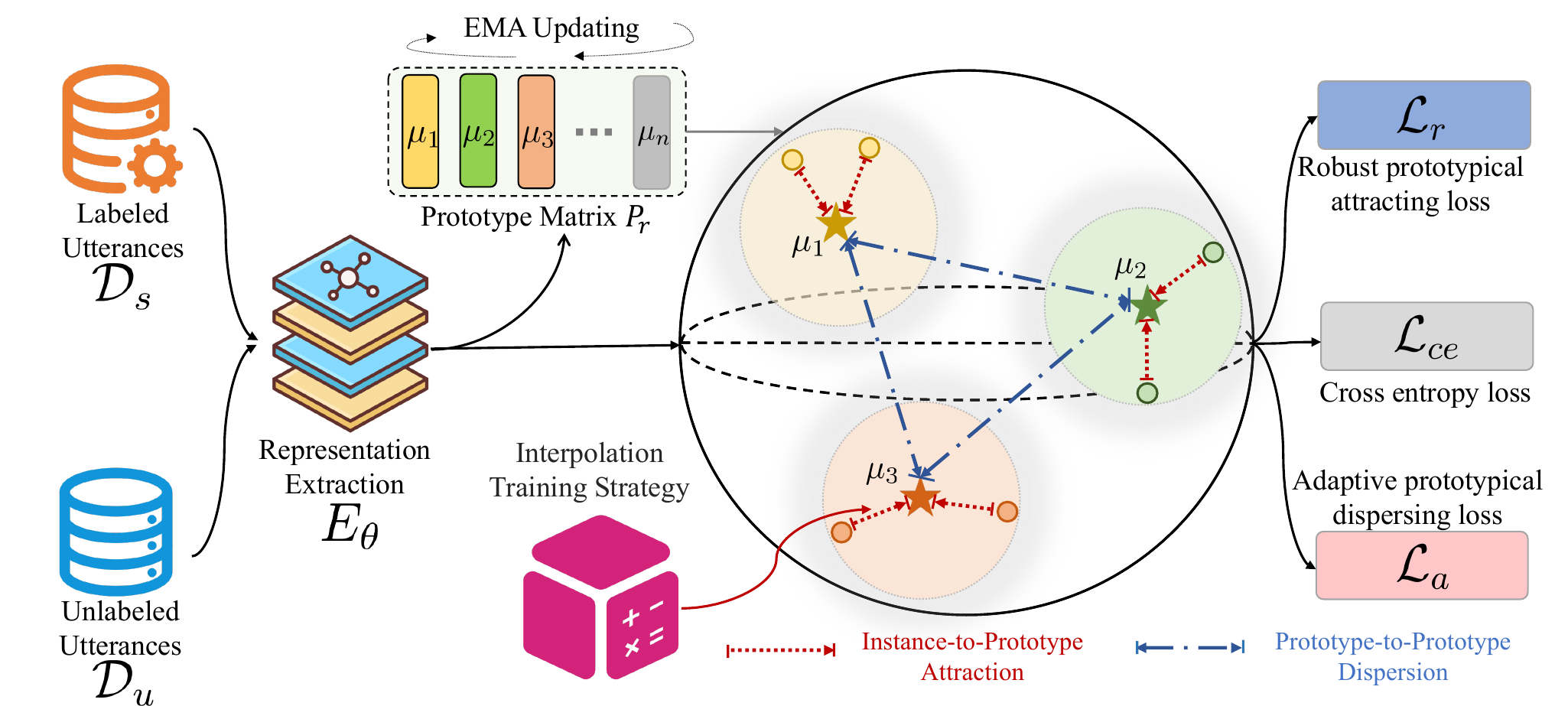}
\caption{
Overview of \ourmethod{}. Our method is jointly optimized by $L_{r}$, $L_{a}$, and $L_{ce}$. $L_{r}$ mitigates the effects of noisy pseudo-labels while minimizes the instance-to-prototype distance, while $L_{a}$ maximizes the prototype-to-prototype distance. $L_{ce}$ is a cross-entropy loss to prevent knowledge forgetting.}
\label{fig:framwork}
\end{figure*}

In this section, we describe the proposed \ourmethod{} framework for new intent discovery in detail. 
As shown in Figure~\ref{fig:framwork}, the architecture of the \ourmethod{} framework contains four main components: 
(1) \textit{Intent representation learning}, which pre-trains a feature extractor $E_{\theta}$ on both labeled and unlabeled intent data to optimize better representation learning (Sec.~\ref{sec:IRL}); 
(2) \textit{Categorical prototypes generation}, which derives the prototypes of the training data by a clustering method (Sec.~\ref{sec:CPG}); 
(3) \textit{Robust prototypical attracting}, which mitigates the effects of noisy pseudo-labels while minimizes the instance-to-prototype distance for stronger within-cluster compactness (Sec.~\ref{sec:RPA});
(4) \textit{Adaptive prototypical dispersing}, which maximizes the prototype-to-prototype distance for larger between-cluster dispersion (Sec.~\ref{sec:APD}).

\subsection{Problem Definition}
New Intent Discovery follows an open-world setting, which aims to recognize all intents with the aid of limited labeled known intent data and unlabeled data containing all classes.
Let $\mathcal{I}_{k}$ and $\mathcal{I}_{n}$ represent the sets of known and novel intents respectively, where $\left\{\mathcal{I}_{k} \cap \mathcal{I}_{n}\right\}=\varnothing$ and $\left|\mathcal{I}_{k}\right|+\left|\mathcal{I}_{n}\right|=C$, where $C$ represents the total number of intent categories.
A typical NID task comprises a set of labeled training set $\mathcal{D}_{s}=\left\{\left(x_{i}, y_{i}\right)\right\}_{i=1}^{N}$, wherein each intent $y_{i} \in \mathcal{I}_{k}$, and a set of unlabeled intent utterances $\mathcal{D}_{u}=\left\{\left(x_{i}\right)\right\}_{i=1}^{M}$, where the intent of each utterance $x_{i}$ belongs to $\left\{\mathcal{I}_{k} \cup \mathcal{I}_{n}\right\}$.
The goal of semi-supervised NID is to use $\mathcal{D}_{s}$ as prior knowledge to help learn clustering-friendly representations to recognize known and discover new intent groups.
After training, the model performance will be evaluated on the testing set $\mathcal{D}_{t} = \{x_{i}|y_{i} \in \mathcal{I}_{k} \cup \mathcal{I}_{n}\}$.


\subsection{Intent Representation Learning}
\label{sec:IRL}
Considering the excellent generalization capability of the pre-trained model, we use the pretrained language model BERT~\cite{Devlin2019BERTPO} as our feature extractor ($E_{\theta}:\mathcal{X} \rightarrow \mathbb{R}^{H})$. 
Firstly, we feed the $i^{th}$ input sentence $x_{i}$ to BERT, and take all token embeddings $[T_0, \dots, T_L]$ $\in$ $\mathds R^{(L+1) \times H}$ from the last hidden layer ($T_0$ is the embedding of the \texttt{[CLS]} token). 
The sentence representation $\boldsymbol{s}_{i} \in \mathbb R^{H}$ is first obtained by applying $\operatorname{mean-pooling}$ operation on the hidden vectors of these tokens:
\begin{MiddleEquation}
\begin{align}
\boldsymbol{s}_{i} = \operatorname{mean-pooling}([\texttt{[CLS]}, T_1,..., T_L])
\end{align} 
\end{MiddleEquation}where $\texttt{[CLS]}$ is the vector for text classification, $L$ is the sequence length, and $H$ is the hidden size.
Motivated by~\cite{zhang2021discovering}, we aim to effectively generalize prior knowledge through pre-training to unlabeled data, we fine-tuned BERT on labeled data ($\mathcal{D}_{s}$) using the cross-entropy (CE) loss.
Furthermore, we follow~\cite{zhang-2022-new-intent-discovery} to use the masked language modeling (MLM) loss on all available data ($\mathcal{D} = \mathcal{D}_{s} \cup \mathcal{D}_{u}$) to learn domain-specific semantics. 
We concurrently pre-train the model with the aforementioned two types of loss:
\begin{MiddleEquation}
\begin{equation}
     \mathcal{L}_{pre} = \mathcal{L}_{ce}(\mathcal{D}_{s}) + \mathcal{L}_{mlm}(\mathcal{D}_{s} \cup \mathcal{D}_{u})
\end{equation}
\end{MiddleEquation}where $\mathcal{D}_{s}$ and $\mathcal{D}_{u}$ are labeled and unlabeled intent corpus, respectively.
The masked language model is trained on the whole corpus $\mathcal{D} = \mathcal{D}_{s} \cup \mathcal{D}_{u}$.
After pretraining, models can acquire diverse general knowledge for both known and novel intents, enabling them to learn meaningful semantic representations for subsequent tasks.

\subsection{Categorical Prototypes Generation}
\label{sec:CPG}
The prototype of an intent class computed as the text average embeddings within the class is defined as a representative embedding for a group of semantically similar instances.
We first obtain the intent embedding $E_{\theta}\left(x_{i}\right)$ for each $x_{i} \in \mathcal{D}_{a}$ and then perform $k$-Means clustering on the training instances $\mathcal{D}_{a}$ to generate $C$ clusters $Q=\{Q_{c}\}_{c=1}^{C}$, where $C$ represents the total number of intent categories. 
The prototype matrix $P_{r}=\{\mu_{c}\}_{c=1}^{C}$ are generated based on their clusters. Concretely, for each cluster $Q_c$, the prototype ${\mu}_{c}$ is computed by:
\begin{MiddleEquation}
\begin{equation}
{\mu}_{c} = \frac{1}{|Q_{c}|} \sum_{x_{i} \in Q_{c}} E_{\theta}\left(x_{i}\right) 
\end{equation}
\end{MiddleEquation}where we presume prior knowledge of $C$ following previous works~\cite{zhang2021discovering} to make a fair comparison and we tackle the problem of estimating this parameter in the experiment (refer to Sec.~\ref{sec:Estimate_C} for a detailed discussion on more accurately estimating $C$ within \ourmethod{}). 

\subsection{Robust Prototypical Attracting}
\label{sec:RPA}
The key idea of the robust prototypical attracting learning (RPAL) method aims to acquire cluster-friendly discriminative representations with strong within-cluster compactness. To minimize the distance between instances and their corresponding prototypes, we employ prototypical contrastive learning (PCL). PCL brings instance representations closer to their matched prototypes while pushing them away from other prototypes as:
\begin{MiddleEquation}
\begin{equation}
\label{eq:4}
\mathcal{L}_{p}=-\mathbb{E}_{i \le N_b}\log \frac{\exp \left(\boldsymbol{z}_{i} \cdot \boldsymbol\mu^{y_{i}} / \tau\right)}{\sum_{k=1}^{C} \exp \left(\boldsymbol{z}_{i} \cdot \boldsymbol\mu^{k} / \tau\right)}
\end{equation}
\end{MiddleEquation}where the normalized sentence embeddings $\boldsymbol{z}_{i}$ matches the prototype $\boldsymbol\mu^{y_{i}}$ of its ground truth label $y_{i}$, $N_b$ is the size of the training set and $\tau$ is a scalar temperature.

\noindent
\textbf{Interpolation Training Strategy} Since the mapping between instances and prototypes derives from pseudo-labels generated by $k$-means clustering, the standard PCL objective is susceptible to noise influence, resulting in suboptimal performance~\cite{jiang2020beyond}. The encoder necessitates a regularization technique to avoid overfitting from forcefully memorizing hard training labels. To further address this issue, we extend the PCL objective by introducing the interpolation training strategy (ITS). By constructing virtual training samples that are linear interpolations of two random samples, the model is forced to predict less confidently on interpolations and produce smoother decision boundaries. Specifically, we first perform convex combinations of instance pairs as:
\begin{MiddleEquation}
\begin{equation}
\label{eq:5}
\boldsymbol{x}^{mix}=\eta \boldsymbol{x}_{a}+(1-\eta) \boldsymbol{x}_{b}
\end{equation}
\end{MiddleEquation}where $\eta \in[0,1] \sim \operatorname{Beta}(\alpha, \alpha)$ and $\boldsymbol{x}^{mix}$ denotes the training sample that combines two samples $\boldsymbol{x}_{a}$ and $\boldsymbol{x}_{b}$, which are randomly chosen from the same minibatch.
We then impose a linear relation in the contrastive loss, which is defined as a weighted combination of the two $\mathcal{L}_{p}$ with respect to class $y_{a}$ and $y_{b}$.
It enforces the embedding for the interpolated input to have the same linear relationship with its corresponding prototypes:
\begin{MiddleEquation}
\begin{gather}
\label{eq:6}
\mathcal{L}_{{r}}=\eta \mathcal{L}_{p}\left({\boldsymbol{z}}^{mix}, \boldsymbol{y}_{a}\right)+(1-\eta) \mathcal{L}_{p}\left({\boldsymbol{z}}^{mix}, \boldsymbol{y}_{b}\right)
\end{gather}
\end{MiddleEquation}where $\boldsymbol{z}^{mix}$ be the normalized embedding for $\boldsymbol{x}^{mix}$, $\boldsymbol{y}_{a}$ and $\boldsymbol{y}_{b}$ are the classes of $\boldsymbol{x}_{a}$ and $\boldsymbol{x}_{b}$.

\subsection{Adaptive Prototypical Dispersing}
\label{sec:APD}
To ensure that the generated intent representations with adequate between-cluster separation to establish distinct cluster boundaries, we draw inspiration from instance-wise contrastive learning:
\begin{MiddleEquation}
\begin{equation}
\label{eq:7}
\begin{aligned}
\mathcal{L}_{c}=
\underbrace{-\frac{1}\tau \mathbb{E}_{i,j} s\left(\boldsymbol{z}_{i}, \boldsymbol{z}_{j}\right)}_{\text{Instance Alignment}}
+\underbrace{\mathbb{E}_{i} \log \sum_{k=1}^{2 N_b} \mathds{1}_{[k \neq i]} e^{s\left(\boldsymbol{z}_{i}, \boldsymbol{z}_{k}\right)}}_{\text {Instance Uniformity}}
\end{aligned}
\end{equation}
\end{MiddleEquation}where the second term is referred to \texttt{uniformity} since it encourages instance representation to be uniformly distributed in the hypersphere. But instance-wise constraints may inevitably lead to the class collision issue~\cite{pmlr-v97-saunshi19a}.


Derived from Eq.~\ref{eq:7}, we devise a novel adaptive prototypical dispersing learning (APDL) method to maximize the prototype-to-prototype distance and improve distribution uniformity by extending the instance-wise contrastive loss. To facilitate large angular distances among different class prototypes, APDL utilizes the distances between prototypes as adaptive weights. This imposes stronger penalties on close prototypes and produces well-separated clusters. The APDL loss is given by:
\begin{MiddleEquation}
\begin{equation}
\label{eq:8}
\begin{aligned}
\mathcal{L}_{a} = 
\underbrace{\mathbb{E}_{i \le C} \log \frac{\sum_{j=1}^{C} \mathds{1}_{[j \neq i]}
{D(\boldsymbol\mu_{i},\boldsymbol\mu_{j})}e^{s(\boldsymbol\mu_{i},\boldsymbol\mu_{j})}}{C-1}}_{\text{Prototypical Uniformity}}
\end{aligned}
\end{equation}
\end{MiddleEquation}where $s(\cdot,\cdot)$ is the cosine similarity to evaluate semantic similarities among prototypes
($s(\boldsymbol\mu_{i},\boldsymbol\mu_{j}) = \cos(\boldsymbol\mu_{i},\boldsymbol\mu_{j})/\tau$).
It is worth noting that $D(\cdot,\cdot)$ is an adaptive constraint term (ACT) that adaptively maximizes the distance between nearer prototypes by taking the reciprocal of their distances:
\begin{MiddleEquation}
\begin{equation}
\label{eq:9}
D(\boldsymbol\mu_{i},\boldsymbol\mu_{j}) = \frac{1} {\left\|\boldsymbol\mu_{i}-\boldsymbol\mu_{j} \right \|_{2} }
\end{equation}
\end{MiddleEquation}where $\boldsymbol\mu_{i}$ and $\boldsymbol\mu_{j}$ represent prototypes of any two intent classes in the intent representation space.

\subsection{Dynamic Prototypes Update}
It is crucial to continuously update the class prototypes over the course of training. 
Although $\boldsymbol{\mu}_k$ can be calculated by averaging the representations of class $k$ across the entire dataset at the end of an epoch, it means the prototypes will remain static during the next full epoch.
This is not ideal, as distributions of intent representations and clusters are rapidly changing, especially in the earlier epochs. So we use the exponential moving average (EMA) algorithm to learn more robust prototypes:
\begin{MiddleEquation}
\begin{equation}
\label{eq:10}
    \begin{split}
        {\boldsymbol{\mu}}_k = \sigma(\lambda{\boldsymbol{\mu}}_k + (1-\lambda) \boldsymbol{z}_{i})      
    \end{split}    
\end{equation}
\end{MiddleEquation}where the $\sigma$ is the layer normalization, $\lambda$ is a momentum factor and $\boldsymbol{z}_{i}$ is normalized embeddings.

\subsection{Multitask Learning}
Our approach learns cluster-friendly intent representations with stronger within-cluster compactness and larger between-cluster separation by jointly performing RPAL and APDL.
To mitigate the risk of catastrophic forgetting of knowledge gained from labeled data, we integrate cross-entropy loss into the training process. The multitask learning objective for NID can be denoted as:
\begin{MiddleEquation}
\begin{equation}
\label{eq:11}
\mathcal{L}_{all} = \omega \mathcal{L}_{r} + \mathcal{L}_{a} + \mathcal{L}_{ce}
\end{equation}
\end{MiddleEquation}where $\omega$ is a hyperparameter that controls the weight of loss. For inference, we perform the non-parametric clustering method $k$-means to obtain cluster assignments for testing data.

\section{Experiments}
\subsection{Datasets}
To validate the effectiveness and generality of our method, we conducted experiments on three diverse and challenging real-world datasets. 
Detailed statistics are described in Table~\ref{tab:stastic_datasets}.
\begin{itemize}
\item \textbf{CLINC}~\cite{larson2019evaluation} is a dataset specially designed for OOD detection and intent discovery, which contains 22.5K samples of user queries in total and 150 unique labeled intents from 10 domains.

\item \textbf{BANKING}~\cite{casanueva2020efficient} is a dataset about banking. The dataset provides user queries and labeled intents from the banking domain, with a total of 13K samples and 77 types of intents.

\item \textbf{StackOverflow}~\cite{xu2015short} is a dataset published in Kaggle.com. It contains 20K samples across 20 classes of technical questions collected from the Kaggle website.
\end{itemize}

\subsection{Baselines}
We compare our approach with unsupervised and semi-supervised models:
\begin{itemize}
\item \textbf{Unsupervised:} 
$k$-means~\cite{macqueen1967some}, 
Agglomerative Clustering (AC)~\cite{gowda1978agglomerative}, 
SAE-KM and Deep Embedded Cluster (DEC)~\cite{xie2016unsupervised}, 
Deep Clustering Network (DCN)~\cite{yang2017towards}, 
DAC~\cite{Chang2017DeepAI}, 
DeepCluster~\cite{caron2018deep}.

\item \textbf{Semi-supervised:} 
PCK-means~\cite{basu2004active}, 
KCL~\cite{hsu2018}, 
MCL~\cite{hsu2019learning},
DTC~\cite{han2019learning},
CDAC+~\cite{lin2020discovering}, 
GCD~\cite{GCD_cvpr2022}, 
DeepAligned~\cite{zhang2021discovering},
MTP-CLNN~\cite{zhang-2022-new-intent-discovery},
ProbNID~\cite{zhou2023latent}, 
DPN~\cite{DPN_AAAI2023}.
\end{itemize}
\begin{table}[t]
\resizebox{0.95\columnwidth}{!}{
    \begin{tabular}{l|c|c|c|c|c}
    \toprule
    Dataset & $|\mathcal{I}_{k}|$ & $|\mathcal{I}_{n}|$ & $|\mathcal{D}_{s}|$ & $|\mathcal{D}_{u}|$ & $|\mathcal{D}_{t}|$ \\
    \midrule
     CLINC & 113 & 37 & 1344 & 16656 & 2250  \\
     BANKING & 58 & 19 & 673 & 8330 & 3080  \\
     StackOverflow & 15 & 5 & 1350 & 16650 & 1000  \\
    \bottomrule
    \end{tabular}}
    \caption{Statistics of datasets. We set the known class ratio $|\mathcal{I}_{k}|/|\mathcal{I}_{k} \cap \mathcal{I}_{n}|$ to $75\%$. The columns represent the number of known categories, novel categories, labeled data, unlabeled data, and testing data, respectively.}
    \label{tab:stastic_datasets}
    \vspace{-5pt}
\end{table}    

\begin{table*}[t]
    \centering
    \resizebox{0.95\textwidth}{!}{
    \begin{tabular}{l l | c c c | c c c | c c c}
    \toprule
     & \multirow{3}{*}{\textbf{Methods}} & \multicolumn{3}{c}{\textbf{CLINC}} & \multicolumn{3}{c}{\textbf{BANKING}} & \multicolumn{3}{c}{\textbf{StackOverflow}} \\
    \cmidrule{3-5} \cmidrule{6-8} \cmidrule{9-11}&  & \textbf{NMI} & \textbf{ARI} & \textbf{ACC} & \textbf{NMI} & \textbf{ARI} & \textbf{ACC} & \textbf{NMI} & \textbf{ARI} & \textbf{ACC} \\
    \midrule
    \multirow{7}*{Unsupervised} & K-means & 70.89 & 26.86 & 45.06 & 54.57 & 12.18 & 29.55 & 8.24 & 1.46 & 13.55 \\
    ~ & AC & 73.07 & 27.70 & 44.03 & 57.07 & 13.31 & 31.58 & 10.62 & 2.12 & 14.66\\
    ~ & SAE-KM & 73.13 & 29.95 & 46.75 & 63.79 & 22.85 & 38.92 & 32.62 & 17.07 & 34.44\\
    ~ & DEC & 74.83 & 27.46 & 46.89 & 67.78 & 27.21 & 41.29 & 10.88 & 3.76 & 13.09 \\
    ~ & DCN & 75.66 & 31.15 & 49.29 & 67.54 & 26.81 & 41.99 & 31.09 & 15.45 & 34.56\\
    ~ & DAC & 78.40 & 40.49 & 55.94 & 47.35 & 14.24 & 27.41 & 14.71 & 2.76 & 16.30 \\
    ~ & DeepCluster & 65.58 & 19.11 & 35.70 & 41.77 & 8.95 & 20.69 & - & - & - \\
    \arrayrulecolor{lightgray}
    \midrule
    \midrule
    \arrayrulecolor{black}
    \multirow{10}*{Semi-supervised} & PCK-Means & 68.70 & 35.40 & 54.61 & 48.22 & 16.24 & 32.66 & 17.26 & 5.35 & 24.16\\
    ~ & KCL (BERT) & 86.82 & 58.79 & 68.86 & 75.21 & 46.72 & 60.15 & 8.84 & 7.81 & 13.94 \\
    ~ & MCL (BERT) & 87.72 & 59.92 & 69.66 & 75.68 & 47.43 & 61.14 & 66.81 & 57.43 & 72.07 \\
    ~ & CDAC+ & 86.65 & 54.33 & 69.89 & 72.25 & 40.97 & 53.83 & 69.84 & 52.59 & 73.48 \\
    ~ & DTC (BERT) & 90.54 & 65.02 & 74.15 & 76.55 & 44.70 & 56.51 & 63.17 & 53.66 & 71.47 \\
    ~ & GCD$^\heartsuit$ & 91.13 & 67.44 & 77.50 & 77.86 & 46.87 & 58.95 & 64.74 & 47.70 & 67.71 \\
    ~ & DeepAligned & 93.95 & 80.33 & 87.29 & 79.91 & 54.34 & 66.59 & 76.47 & 62.52 & 80.26 \\
    ~ & MTP-CLNN$^\heartsuit$ & 94.88 & 84.77 & 88.25 & 84.22 & 63.10 & 73.98 & 77.03 & 69.50 & 83.18 \\
    ~ & ProbNID & 95.01 & 83.00 & 88.99 & 84.02 & 62.92 & 74.03 & 77.32 & 65.70 & 80.50 \\
    ~ &  DPN$^\heartsuit$ & 95.14 & 84.30 & 89.22 & 84.31 & 63.26 & 74.45 &  79.89 & 70.27 & 84.59 \\
    \midrule
    ~ & \textit{\ourmethod{}(Ours)} & \textbf{\underline{95.93}} & \textbf{\underline{86.28}} & \textbf{\underline{91.24}}& \textbf{\underline{85.16}} & \textbf{\underline{65.79}} & \textbf{\underline{76.27}} & \textbf{\underline{82.36}} & \textbf{\underline{71.73}} & \textbf{\underline{86.60}} \\
    \bottomrule
    \end{tabular}
    }
    \caption{Comparison against the unsupervised and semi-supervised baselines on three benchmarks.
    $\heartsuit$ denotes results obtained from running the provided code and other results are retrieved from~\citet{zhou2023latent}. Results are averaged over different random seeds and the bold fonts denote the best scores.}
    \label{tab:Main_results}
\end{table*}

\subsection{Evaluation Metrics}
To evaluate the quality of the discovered intent clusters, we use three broadly used evaluation metrics~\cite{zhang2021discovering,zhang-2022-new-intent-discovery,zhou2023latent}:
(1) Normalized Mutual Information \textbf{(NMI)} measures the normalized mutual dependence between the predicted labels and the ground-truth labels.
(2) Adjusted Rand Index \textbf{(ARI)} measures how many samples are assigned properly to different clusters.
(3) Accuracy \textbf{(ACC)} is measured by assigning dominant class labels to each cluster and taking the average precision.
Higher values of these metrics indicate better performance.
Specifically, NMI is defined as:
\begin{MiddleEquation}
\begin{align}
    \textrm{NMI}(\mathbf{y}^{gt}, \mathbf{y}^{p}) &=  \frac{\it{MI}(\mathbf{y}^{gt}, \mathbf{y}^{p})}{\frac{1}{2} (H(\mathbf{y}^{gt}) + H(\mathbf{y}^{p}))},
\end{align}
\end{MiddleEquation}where $\mathbf{y}^{gt}$ and $\mathbf{y}^{p}$  are the ground-truth  and predicted labels, respectively. $\it{MI}(\mathbf{y}^{gt}, \mathbf{y}^{p})$ represents the mutual information between $\mathbf{y}^{gt}$ and $\mathbf{y}^{p}$, and $H(\cdot)$ is the entropy.  $\it{MI}(\mathbf{y}^{gt}, \mathbf{y}^{p})$ is normalized by the arithmetic mean of $H(\mathbf{y}^{gt})$ and $H(\mathbf{y}^{p})$, and the values of $\textrm{NMI}$ are in the range of [0, 1]. ARI is defined as:
\begin{MiddleEquation}
\begin{align}
    \textrm{ARI} &= \frac{
    \sum_{i, j}\binom{n_{i, j}}{2}-[\sum_{i}\binom{u_{i}}{2}\sum_{j}\binom{v_{j}}{2}] / \binom{n}{2}
    }
    {
    \frac{1}{2}[\sum_{i}\binom{u_{i}}{2}+\sum_{j}\binom{v_{j}}{2}]-[\sum_{i}\binom{u_{i}}{2}\sum_{j}\binom{v_{j}}{2}]/\binom{n}{2}
    }
\end{align}
\end{MiddleEquation}where $u_{i}=\sum_{j}n_{i,j}$, and $v_{j}=\sum_{i}n_{i,j}$. $n$ is the number of samples, and $n_{i,j}$ is the number of the samples that have both the $i^{\textrm{th}}$ predicted label  and the $j^{\textrm{th}}$ ground-truth label. The values of ARI are in the range of [-1, 1]. ACC is defined as:
\begin{MiddleEquation}
\begin{align}
     \textrm{ACC}(\mathbf{y}^{gt}, \mathbf{y}^{p}) &=\max _m \frac{\sum_{i=1}^n \mathbb{I} \left\{y^{gt}_i=m\left(y^{p}_i\right)\right\}}{n}
\end{align}
\end{MiddleEquation}where $m$ is a one-to-one mapping between the ground-truth label $\mathbf{y}^{gt}$ and predicted label $\mathbf{y}^{p}$ of the $i^{\textrm{th}}$ sample. The Hungarian algorithm is used to obtain the best mapping $m$ efficiently. The values of ACC are in the range of [0, 1].


\subsection{Experimental Settings}
In experiments, we utilize the pre-trained 12-layer bert-uncased BERT model\footnote{\url{https://huggingface.co/bert-base-uncased}} \cite{Devlin2019BERTPO} as the backbone encoder and only fine-tune the last transformer layer parameters to expedite the training process. 
For model optimization, we adopt the AdamW~\cite{adamW} optimizer with a learning of 1e-5.
The wait patience for early stopping is set to 20.
For masked language modeling, the mask probability is set to 0.15 following previous work.
For MTP-CLNN, the external dataset is not used as in other baselines, the parameter of top-$k$ nearest neighbors is set to $\{$100, 50, 500$\}$ for CLINC, BANKING, and StackOverflow, respectively, as utilized in~\citet{zhang-2022-new-intent-discovery}. 
For all experiments, we split the datasets into train, valid, and test sets, and randomly select 25\% of categories as unknown and only 10\% of training data as labeled~\cite{zhang2021discovering}.
The number of intent categories is set as ground truth. 
we set the temperature scale as $\tau$ = 0.1 in Eq.~\eqref{eq:4} and Eq.~\eqref{eq:8}, the parameter in beta distribution $\alpha$ = 1 in Eq.~\eqref{eq:5} (i.e. $\eta$ is sampled from a uniform distribution), the momentum factor $\lambda$ = 0.9 in Eq.~\eqref{eq:10}.
All the experiments are conducted on 4 Tesla V100 GPUs and averaged over 10 different seeds.

\begin{table*}[t]
    \centering
    \resizebox{0.8\textwidth}{!}{
    \begin{tabular}{l | c c c | c c c | c c c }
    \toprule
     \multirow{3}{*}{\textbf{Methods}} & \multicolumn{3}{c}{\textbf{CLINC}} & \multicolumn{3}{c}{\textbf{BANKING}} & \multicolumn{3}{c}{\textbf{StackOverflow}}\\
     \cmidrule{2-4} \cmidrule{5-7}  \cmidrule{8-10}  
    ~ & \textbf{NMI} & \textbf{ARI} & \textbf{ACC} & \textbf{NMI} & \textbf{ARI} & \textbf{ACC} & \textbf{NMI} & \textbf{ARI} & \textbf{ACC}\\
    \midrule
        \textit{\ourmethod{}} & \textbf{\underline{95.93}} & \textbf{\underline{86.28}} & \textbf{\underline{91.24}} & \textbf{\underline{85.16}} & \textbf{\underline{65.79}} & \textbf{\underline{76.27}} & \textbf{\underline{82.36}} & \textbf{\underline{71.73}} & \textbf{\underline{86.60}} \\
    \midrule
    {\large{\ding{172}}} w/o $\mathcal{L}_{{rpal}}$ & 93.77 & 81.42 & 87.05 & 80.82 & 59.15 & 71.33 & 76.06 & 66.55 & 80.72 \\
    {\large{\ding{173}}} w/o $\mathcal{L}_{{apdl}}$ & 93.13 & 82.58 & 88.39 & 82.37 & 62.03 & 72.15 & 79.15 & 69.07 & 83.67 \\
    {\large{\ding{174}}} w/o $\mathcal{L}_{{ce}}$ & 95.02 & 85.27 & 90.17 & 84.49 & 65.27 & 75.00 & 81.18 & 71.06 & 85.44 \\
    {\large{\ding{175}}} w/o ITS & 94.93 & 85.09 & 90.90 & 83.72 & 63.18 & 74.96 &  81.77 & 70.50 & 85.21 \\
    {\large{\ding{176}}} w/o ACT & 95.49 & 85.70 & 89.84 & 84.58 & 63.91 & 75.77 & 82.19 & 70.44 & 86.34 \\
    \bottomrule
    \end{tabular}}
    \caption{Experimental results of ablation study on the CLINC, BANKING, and StackOverflow datasets.}
    \label{tab:ablation}
\end{table*}

\subsection{Main Results}
Table~\ref{tab:Main_results} shows the main results on three datasets.
It is observed that \ourmethod{} achieves the overall best performances compared to other baselines across all datasets.
The in-depth observations can be derived from the results:
(1) Compared with unsupervised methods (Unsupervised), semi-supervised methods (Semi-supervised) achieve much better results, which demonstrates the advantage of prior knowledge transfer for subsequent tasks.
(2) Under the semi-supervised setting (Semi-supervised), our method achieves new state-of-the-art results across all datasets and metrics. 
The effectiveness of our method can be attributed to its ability to efficiently control and coordinate both within-cluster and between-cluster distances, even with limited labeled data as prior knowledge. 
It is helpful to establish distinct global decision boundaries between known and novel intent categories, thereby excelling in NID.
We also find our method yields competitive results on imbalanced datasets like BANKING, underscoring the robust generalization capabilities of our model.
This indicates that our method is better tailored for real-world NID tasks.


\subsection{Ablation Study}
To investigate the contribution of each component within \ourmethod{}, we conduct an ablation study across all datasets in Table~\ref{tab:ablation} and consider five sub-modules as variants. Removing any component will lead to performance degradation, emphasizing the essence of each independent component.
Specifically,
(1) \ourmethod{} (\texttt{w/o} $\mathcal{L}_{{rpal}}$) refers to removing the robust prototypical attracting learning objective.
Experiment {\large{\ding{172}}} indicates that our method is adept at minimizing the instance-to-prototype distances, thereby enhancing within-cluster compactness.
(2) \ourmethod{} (\texttt{w/o} $\mathcal{L}_{{apdl}}$) involves removing adaptive prototypical dispersing learning objective.
Experiment {\large{\ding{173}}} suggests that augmenting between-cluster dispersion is pivotal for optimizing NID performance. Without explicitly constraining prototype-to-prototype distances, prior methods hinder the model from acquiring cluster-friendly representations.
(3) \ourmethod{} (\texttt{w/o} $\mathcal{L}_{{ce}}$) indicates the exclusion of the cross-entropy loss term during the joint optimization process for our method.
Experiment {\large{\ding{174}}} highlights the importance of $\mathcal{L}_{{ce}}$ in mitigating catastrophic forgetting of knowledge learned from labeled data.
(4) \ourmethod{} (\texttt{w/o} ITS) refers to removing the interpolation training strategy in Eq.~\eqref{eq:6}, which implies substituting Eq.\eqref{eq:4} in place of Eq.~\eqref{eq:6}.
Experiment {\large{\ding{175}}} connotes the efficacy of the ITS in mitigating pseudo-label noise.
(5) \ourmethod{} (\texttt{w/o} ACT) refers to removing the adaptive constraint term $\text{dist}(\cdot,\cdot)$ in Eq.~\eqref{eq:8}.
Experiment {\large{\ding{176}}} verifies the importance of imposing stricter penalties on nearer prototypes to optimize the between-cluster distances.

\begin{figure*}[t]
    \centering
    \subfigure[Strong Baseline]{
    \includegraphics[width=0.47\columnwidth]{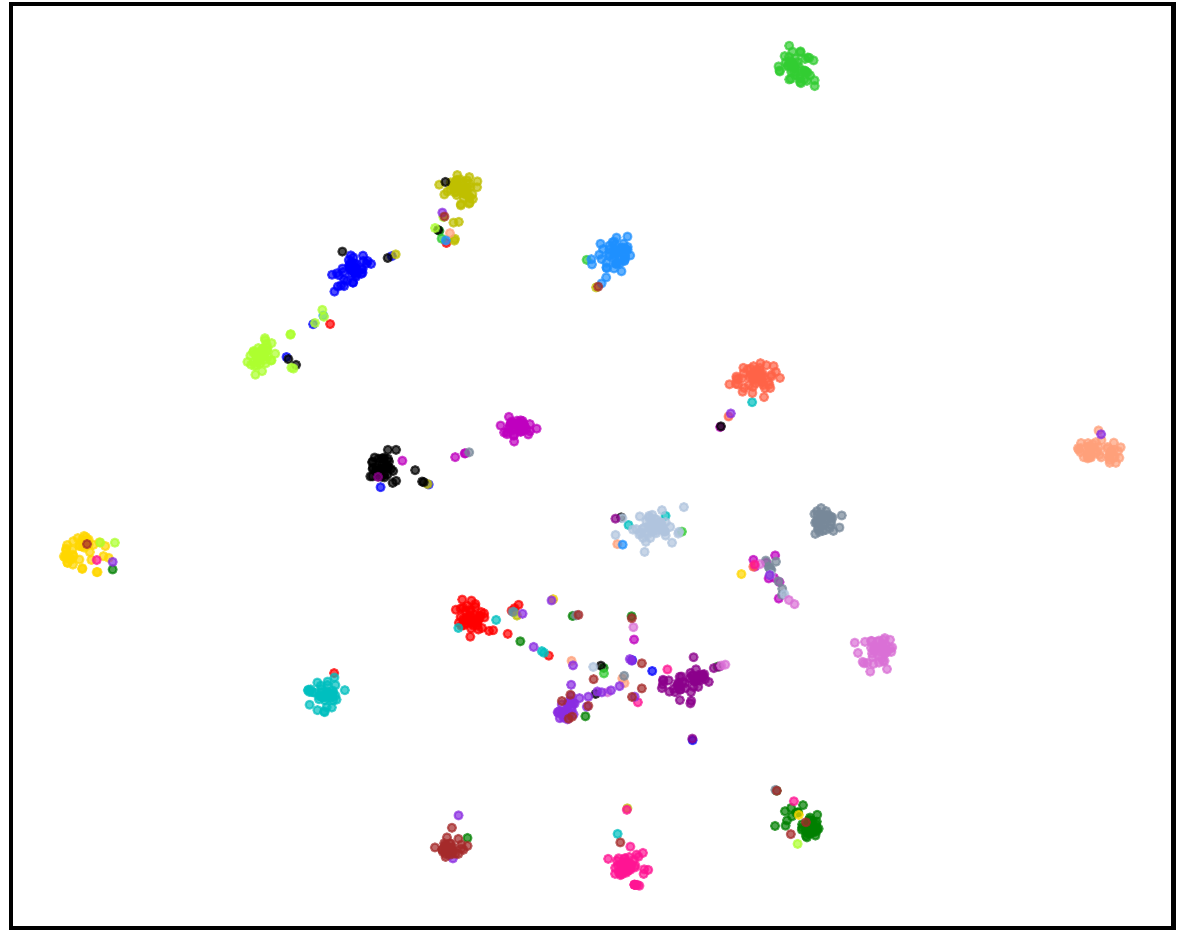}
    \label{tsne_1}
    }
    \subfigure[\ourmethod{} w/o RPAL (Ours)]{
    \includegraphics[width=0.47\columnwidth]{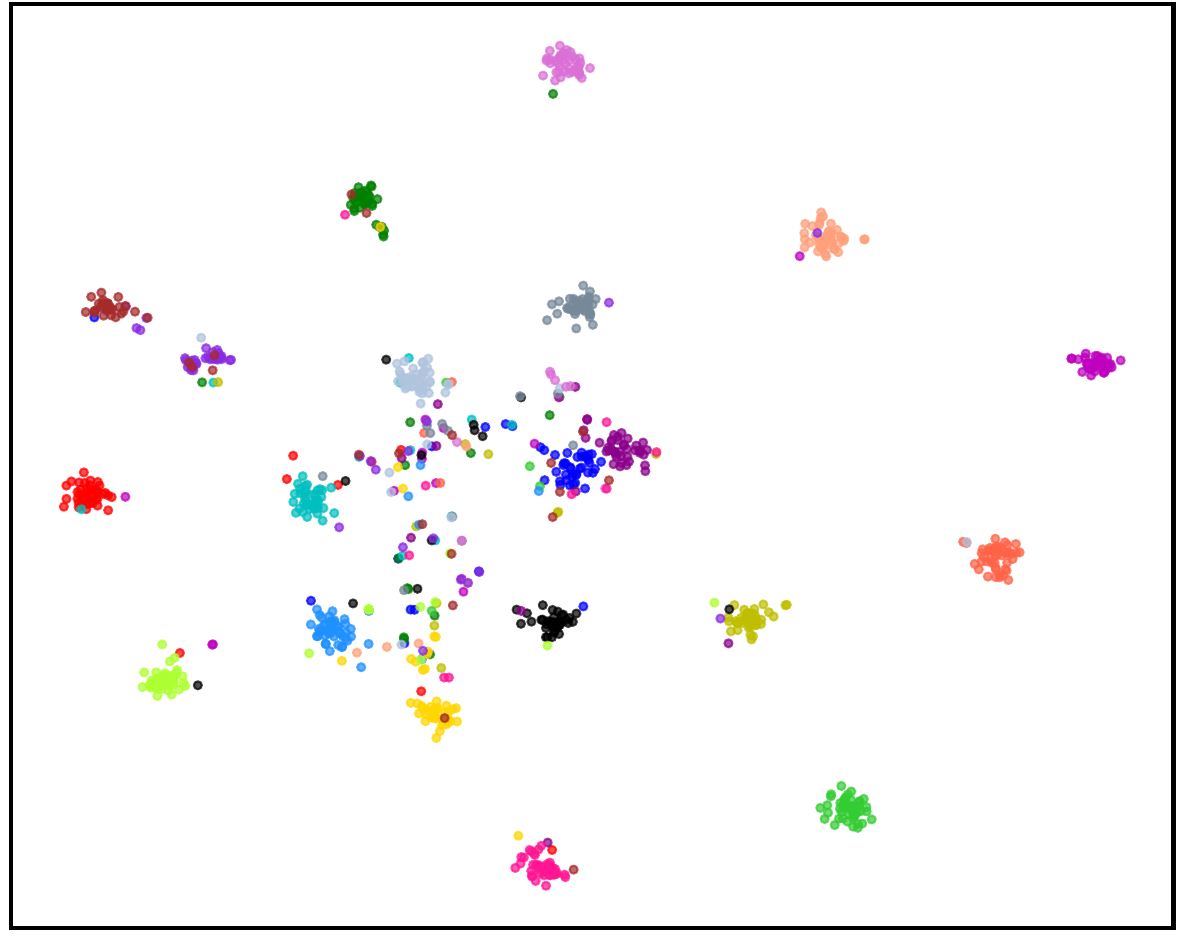}
    \label{tsne_2}
    }
    \subfigure[\ourmethod{} w/o APDL (Ours)]{
    \includegraphics[width=0.47\columnwidth]{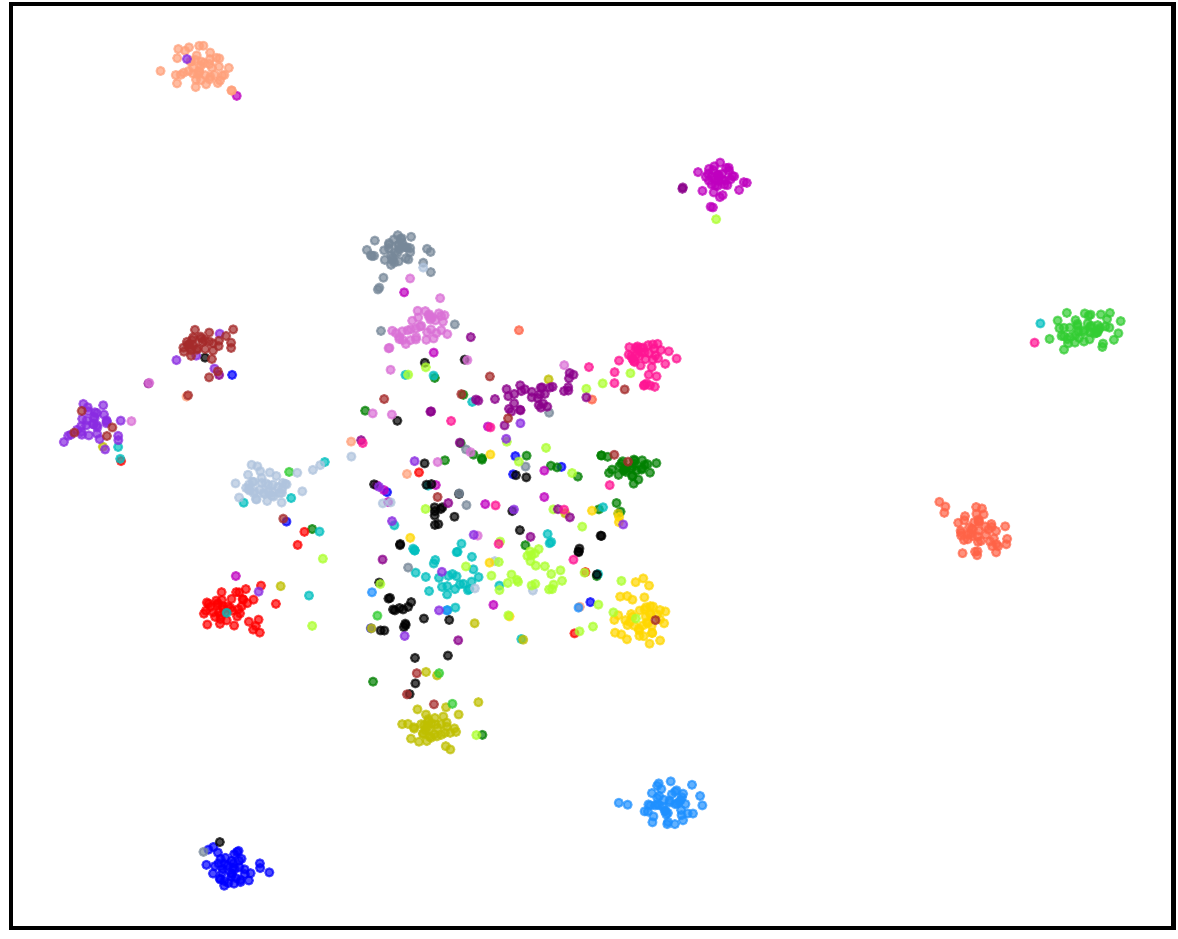}
    \label{tsne_3}
    }
    \subfigure[\ourmethod{} (Ours)]{
    \includegraphics[width=0.47\columnwidth]{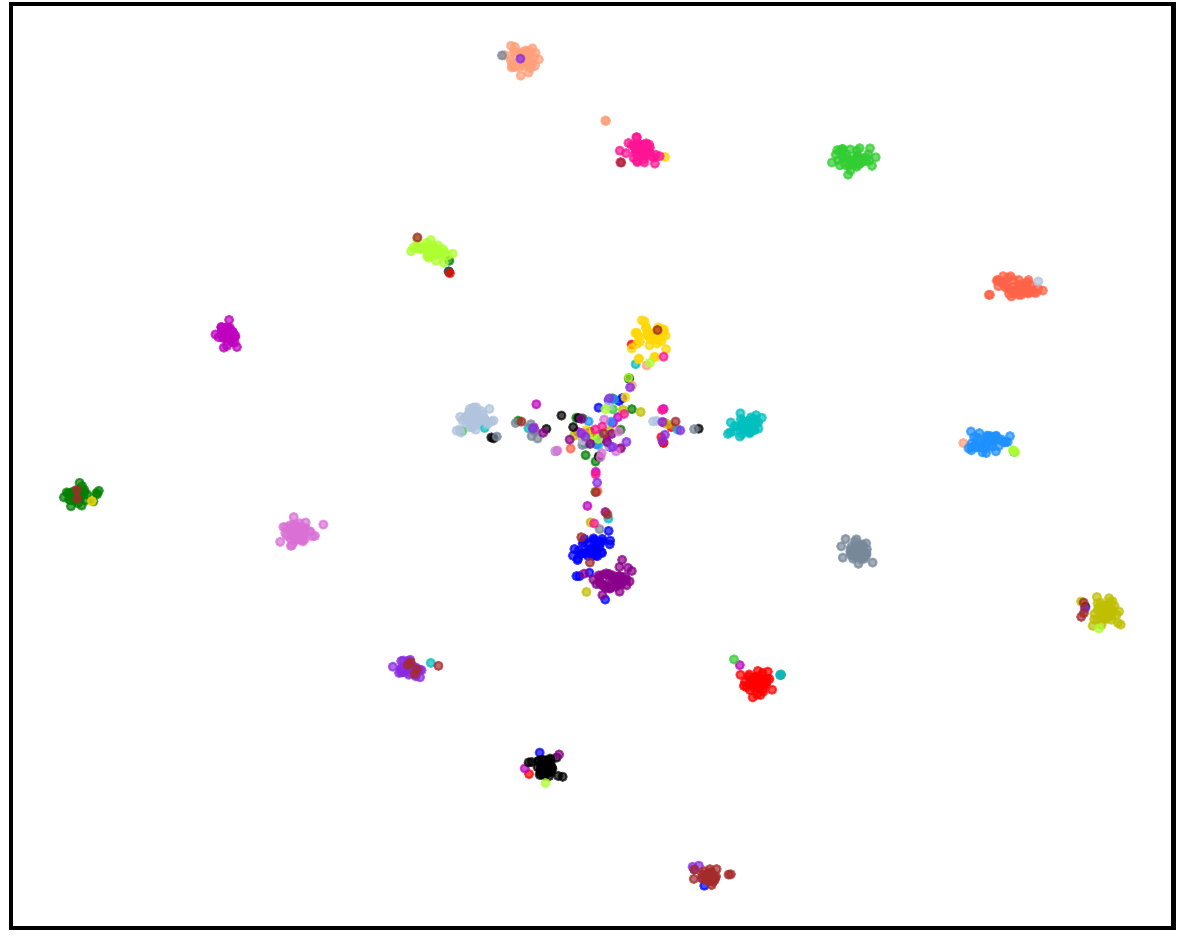}
    \label{tsne_4}
    }
    \caption{t-SNE visualization of learned representation.} 
    \label{fig:t-SNE}
\end{figure*}

\begin{figure*}[t]
    \centering
    \subfigure[Effect on CLINC]{
    \includegraphics[width=0.5\columnwidth]{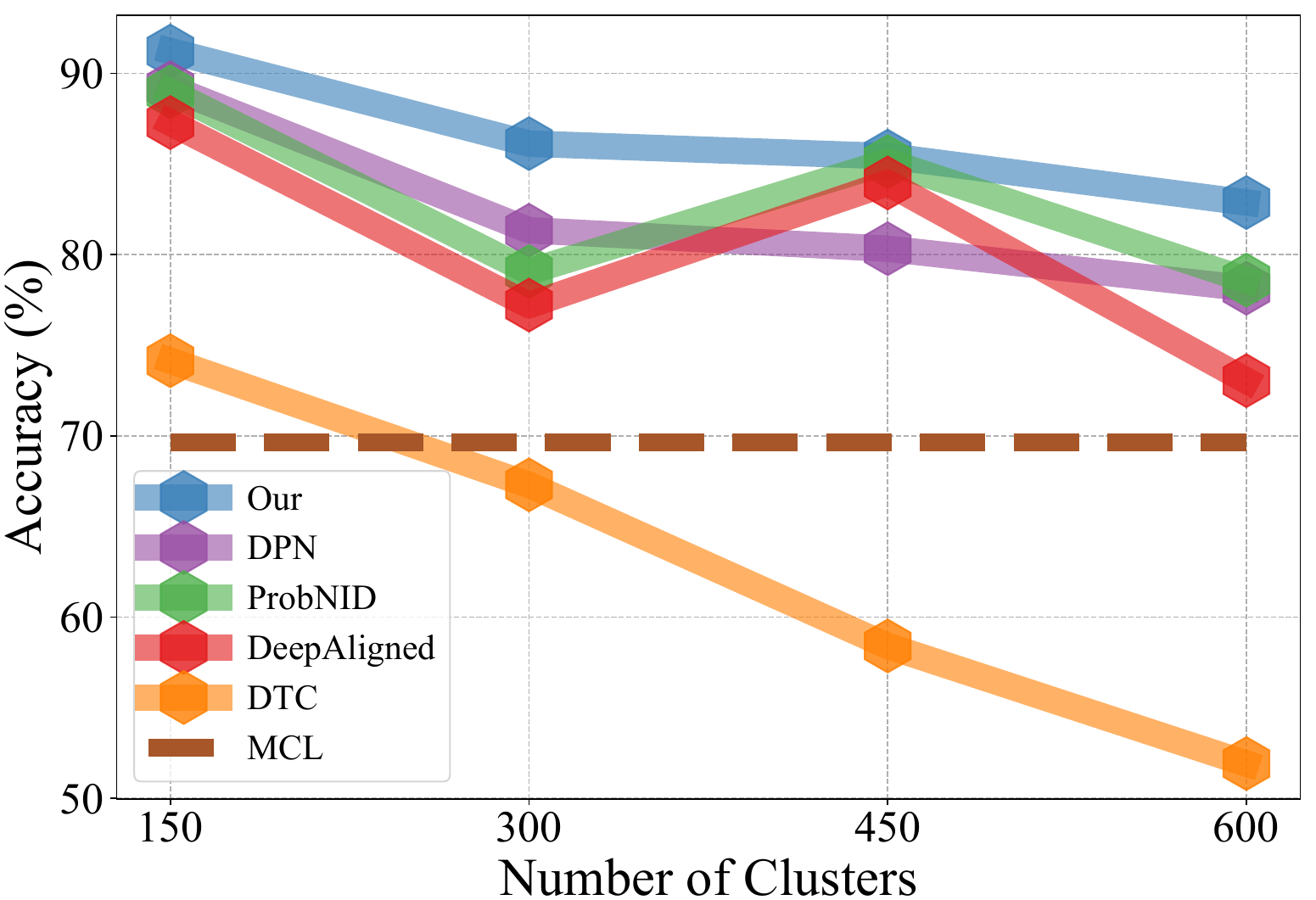}
    \label{subfigure: initial_clusters_clinc}
    }
    \subfigure[Effect on BANKING]{
    \includegraphics[width=0.5\columnwidth]{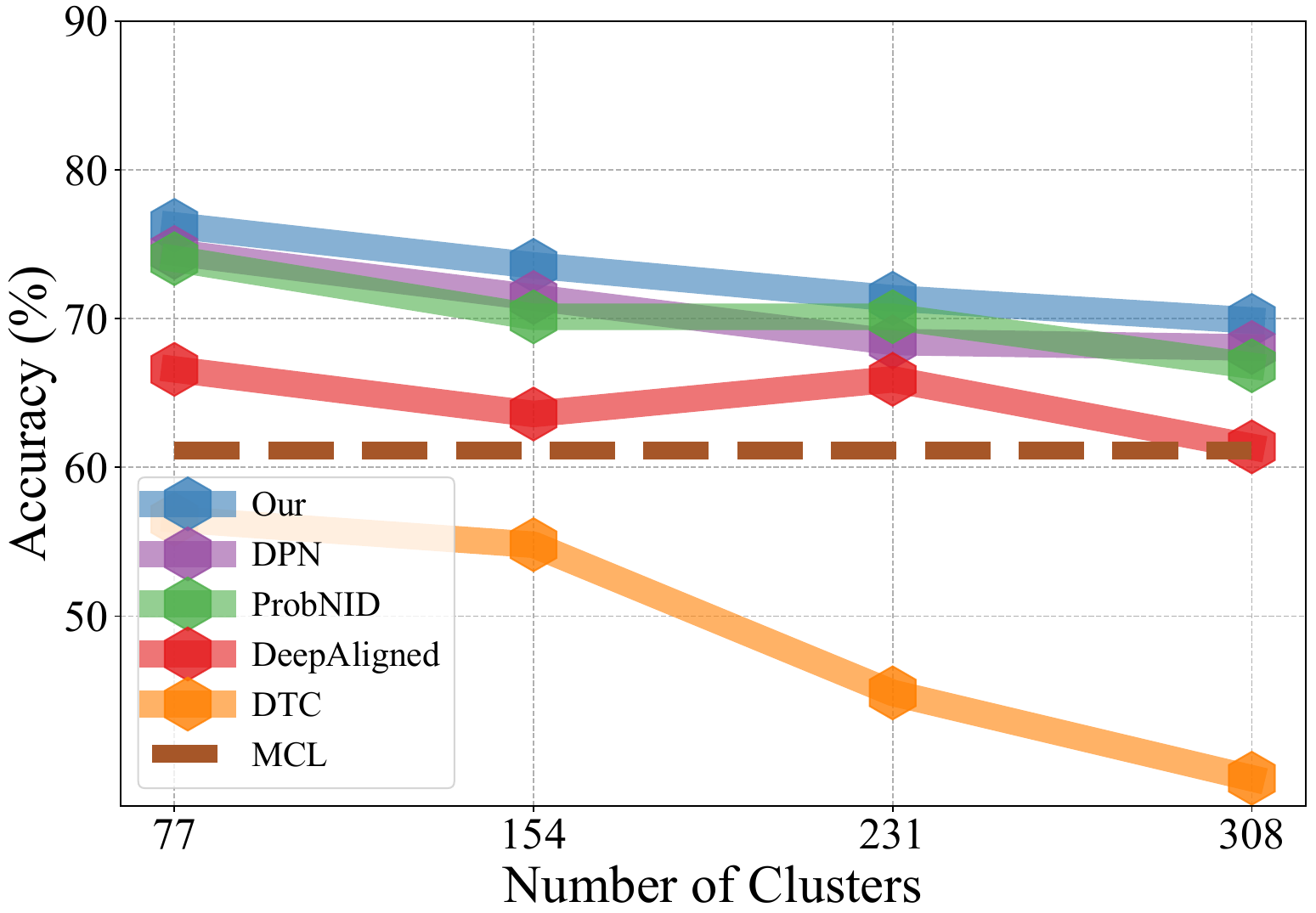}
    \label{subfigure: initial_clusters_banking}
    }
    \subfigure[Effect on StackOverflow]{
    \includegraphics[width=0.5\columnwidth]{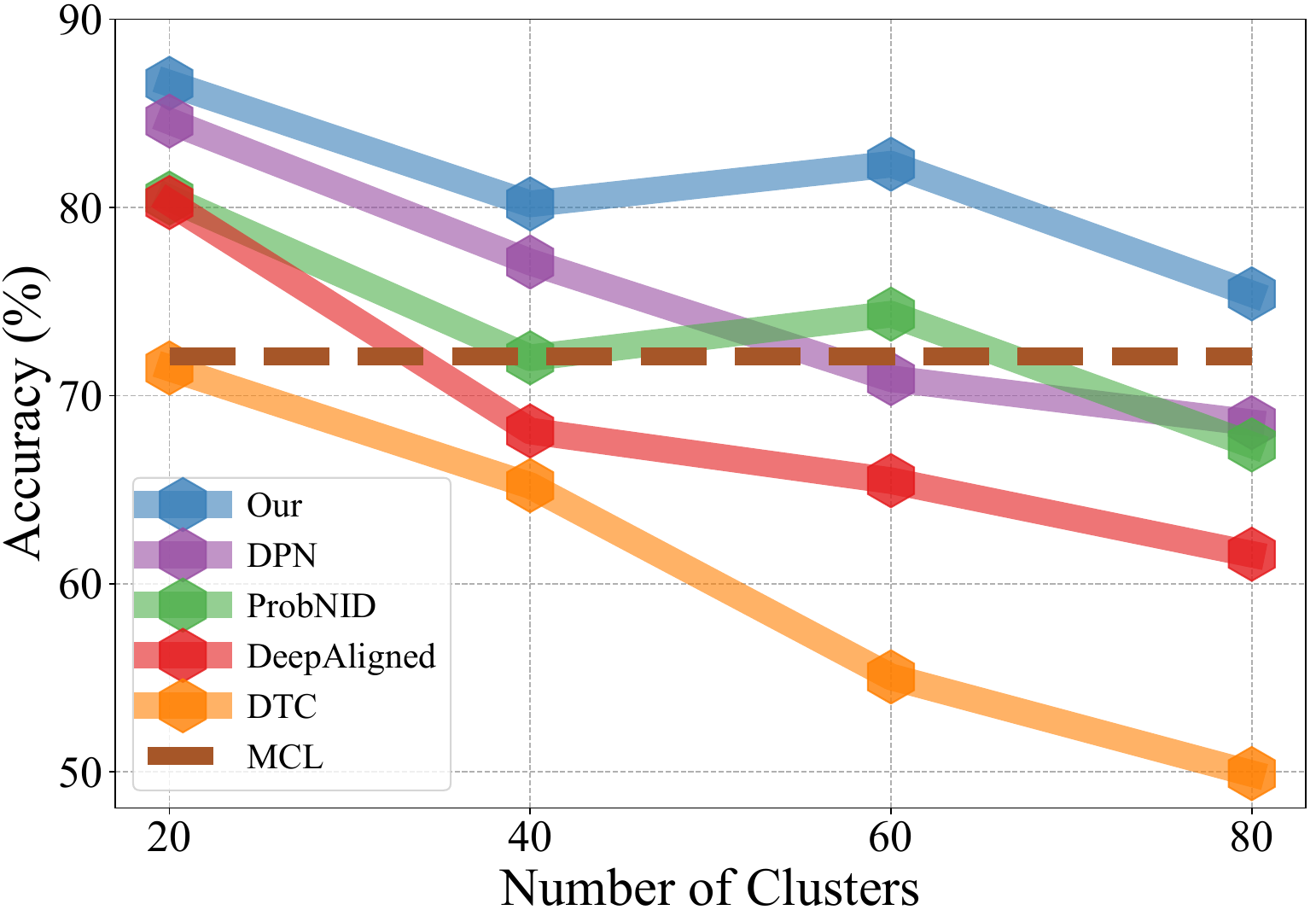}
    \label{subfigure: initial_clusters_stack}
    }
    \caption{Sensitivity of the models to the number of initial clusters on three datasets.}
    \label{fig:initial_clusters}
\end{figure*}

\section{Discussion}
\subsection{Compactness and Separability}
\ourmethod{} aims to generate cluster-friendly representations with two vital characteristics: Compactness and Separability.
We investigate these two properties from the following perspectives.

\noindent
\textbf{Stronger within-cluster compactness.}
To show the capability of our method in promoting tight within-cluster representations, we measure the within-cluster distance by average cosine similarity calculation between each intent embedding and its corresponding prototype in Table~\ref{tab: Compactness_Separability}.
We can see that our method achieves a significantly lower within-cluster distance compared to previous leading methods. This phenomenon may be attributed to the pivotal role of RPAL in enhancing within-cluster compactness.

\noindent
\textbf{Larger between-cluster dispersion.}
To further analyze whether our method truly enlarges the distances among prototypes, we compute the mean cosine similarity for all pairs of class prototypes.
In Table~\ref{tab: Compactness_Separability}, the proposed \ourmethod{} consistently obtains larger between-cluster distances compared to previous competitive methods. 
We speculate that the primary reason for this finding is that APDL plays a crucial role in enlarging the between-cluster dispersion.
This also fully conforms to our expectations that the APDL effectively improves the uniformity of the intent representation space.

\begin{table}[t]
    \centering
    \resizebox{0.95\columnwidth}{!}{
    \begin{tabular}{l | c | c }
    \toprule
     \textbf{Methods} & \textbf{Within} ($\downarrow$) & \textbf{Between} ($\uparrow$) \\
    \midrule
    DTC (BERT) & 22.40  &  11.82 \\
    DeepAligned & 11.75 & 13.91 \\ 
    DPN & 9.03 & 17.38 \\ 
    \textit{Ours} \texttt{(CLINC)} & \textbf{\underline{8.68}} & \textbf{\underline{23.36}}\\
    \arrayrulecolor{lightgray}
    \midrule
    \midrule
    \arrayrulecolor{black}
    DTC (BERT) & 20.31  & 22.49  \\
    DeepAligned & 10.28 & 13.14 \\ 
    DPN & 7.96 & 11.08 \\ 
    \textit{Ours} \texttt{(BANKING)} & \textbf{\underline{7.22}} & \textbf{\underline{17.50}}\\
    \arrayrulecolor{lightgray}
    \midrule
    \midrule
    \arrayrulecolor{black}
    DTC (BERT) & 17.11  & 7.98  \\
    DeepAligned & 9.73 & 11.53 \\ 
    DPN & 7.30 & 14.66  \\ 
    \textit{Ours} \texttt{(StackOverflow)} & \textbf{\underline{4.07}} & \textbf{\underline{20.81}}\\
    \bottomrule
    \end{tabular}
    }
    \caption{Statistics of within-cluster and between-cluster distances~\cite{islam2021broad}.}
    \vspace{-5pt}
    \label{tab: Compactness_Separability}
\end{table}

\subsection{Representation Visualization}
To further validate the effectiveness of our method in learning discriminative intent representations, we adopt the t-SNE to visualize projected representation on the StackOverflow dataset.
Comparing Figure~\ref{tsne_1} and Figure~\ref{tsne_4}, we can clearly see that the clusters obtained by our method are generally more compact and well-separated than those obtained by the strong baseline model. This indicates that our model learns cluster-friendly features for NID.
Looking at Figure~\ref{tsne_2} and Figure~\ref{tsne_4}, it evidently shows that the RPAL effectively pulls instances closer to their corresponding prototypes, achieving strong within-cluster compactness.
Moreover, the difference between Figure~\ref{tsne_3} and Figure~\ref{tsne_4} shows that the APDL significantly pushes prototypes away from each other and builds distinct cluster boundaries.

\subsection{Effect of the Number of Clusters}
To explore the sensitivity of the models to the initial number of clusters $C$, we adjust $C$ from its ground-truth value up to four times that amount.
In Figure~\ref{fig:initial_clusters}, we observe that most methods show a performance drop with an increasing initial value of $C$.
This is because the unreasonably assigned $C$ leads to the generation of noisy pseudo-labels, which substantially impacts the clustering results.
Notably, our method still achieves the best performance on three datasets, validating the capacity of our model to mitigate the impact of noisy pseudo-labels and augment robustness.
This also shows the superiority of our method in the same semi-supervised NID setting.

\begin{table}[t]
    \centering
    \resizebox{0.95\columnwidth}{!}{
    \begin{tabular}{l | c c  | c c | c c}
    \toprule
     \multirow{3}{*}{\textbf{Methods}} & \multicolumn{2}{c}{\textbf{CLINC}} & \multicolumn{2}{c}{\textbf{BANKING}} & \multicolumn{2}{c}{\textbf{StackOverflow}} \\
    \cmidrule{2-3} \cmidrule{4-5} \cmidrule{6-7} 
    ~ & $C$ & $E$  & $C$ & $E$ & $C$ & $E$ \\
    \midrule
    DTC (BERT) & 112 & 25.33  & 58 & 24.68 & 26 & 30.00 \\
    DeepAligned & 129 & 14.00  & 67 & 12.99& 17 & 15.00 \\
    ProbNID & 130 & 13.30 & 73 & 5.48 & - & - \\ 
    DPN & 137 & 8.67 & 71 & 7.80 & 18 & 10.00 \\ 
    \midrule
    \textit{Ours} & \textbf{\underline{141}} & \textbf{\underline{6.00}} & \textbf{\underline{75}} & \textbf{\underline{2.60}} & \textbf{\underline{22}} & \textbf{\underline{10.00}}\\
    \bottomrule
    \end{tabular}
    }
    \caption{Estimation of the number of clusters $C$, where $E$ represents the error rate, which is obtained by calculating the estimated $C$ and the ground truth number.}
    \label{tab: Estimate}
\end{table}

\subsection{Estimate Number of Clusters}
\label{sec:Estimate_C}
The above experiments assume the number of clusters $C$ to be the ground truth. But this is unrealistic in practice.
Therefore, in order to further validate the effectiveness of our method in practical scenarios, we conduct experiments to estimate the number of clusters.
We use the same settings as ~\citet{zhang2021discovering}  and firstly assign the number of intents as two times the ground truth number to investigate the ability to estimate $C$.
In Table~\ref{tab: Estimate}, we notice that our method can predict the number of intents more accurately and achieve better results at the same time. 
The results indicate that our method more easily learns cluster-friendly discriminative representations that assist in accurately estimating the number of clusters.

\begin{figure}[t]
    \centering
    \subfigure[Impact on CLINC]{
    \includegraphics[width=0.45\columnwidth]{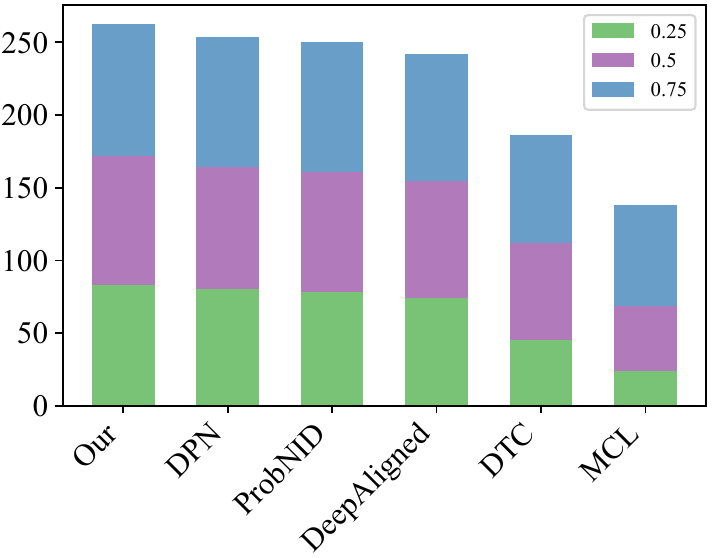}
    \label{subfigure:ratio_clinc}
    }
    \subfigure[Impact on BANKING]{
    \includegraphics[width=0.45\columnwidth]{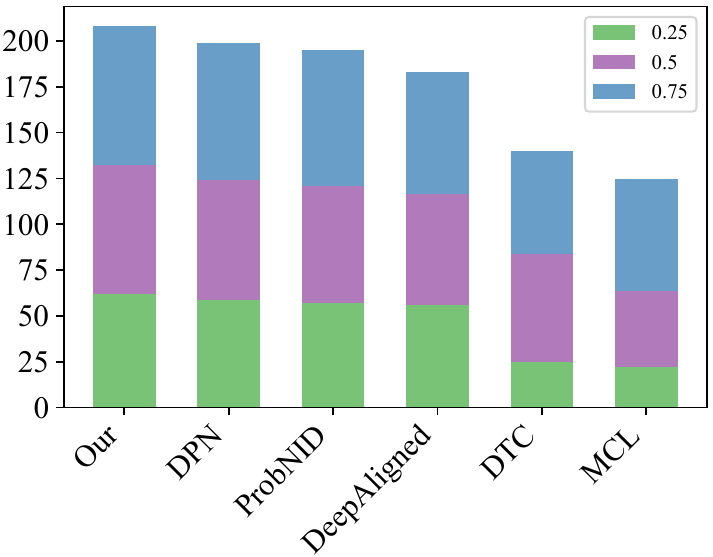}
    \label{subfigure:ratio_banking}
    }
    \vspace{-10pt}
    \caption{Impact of varying the known class ratio on two datasets. The x-axis represents different models and the y-axis denotes their corresponding accuracy values.}
    \vspace{-10pt}
    \label{fig:fig_radio_}
\end{figure}

\subsection{Effect of Known Class Ratio}
To investigate the influence of the number of known intents, we vary the known class ratio in the range of 25\%, 50\% and 75\% during training.
From Figure~\ref{fig:fig_radio_}, it is evident that the performance of all strong methods gradually decreases as the known intent rate decreases. 
As the known intent rate decreases, there is less labeled data available to guide model training, which complicates the transfer of prior knowledge for discovering new intents. 
However, with the decrease in the known intent rate, our proposed \ourmethod{} demonstrates more significant improvements.
We surmise that as the number of intent categories increases, the pivotal factor for enhancing performance is the learning of cluster-friendly representations, which establish distinct boundaries for both known and novel categories.
This highlights the proficiency of our model in optimizing both within-cluster and between-cluster distances, resulting in well-defined cluster boundaries.


\begin{figure}[t]
    \centering
    \subfigure[Effect on ACC]{
    \includegraphics[width=0.45\columnwidth]{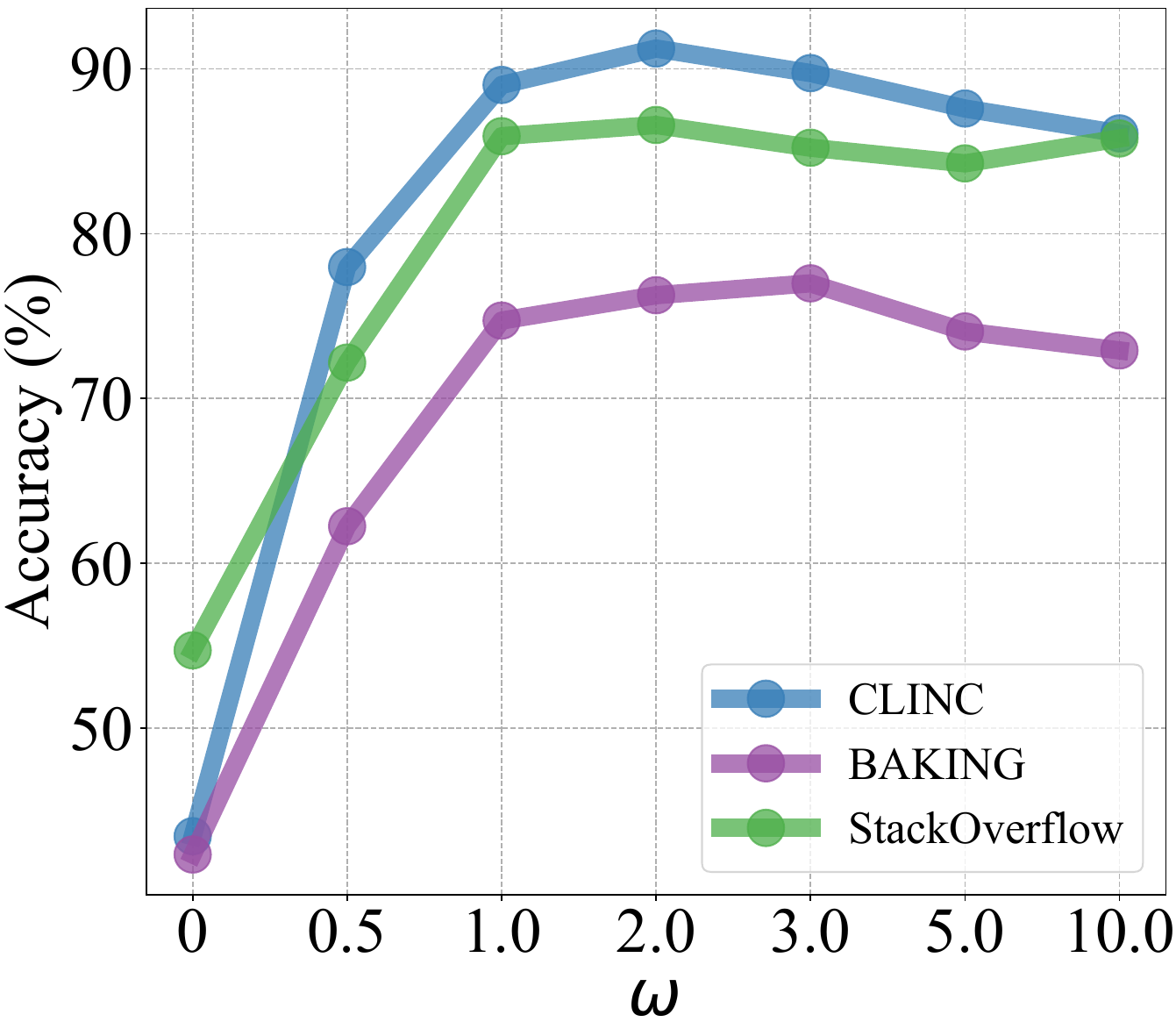}
    \label{subfigure: scatter_bert2}
    }
    \subfigure[Effect on NMI]{
    \includegraphics[width=0.45\columnwidth]{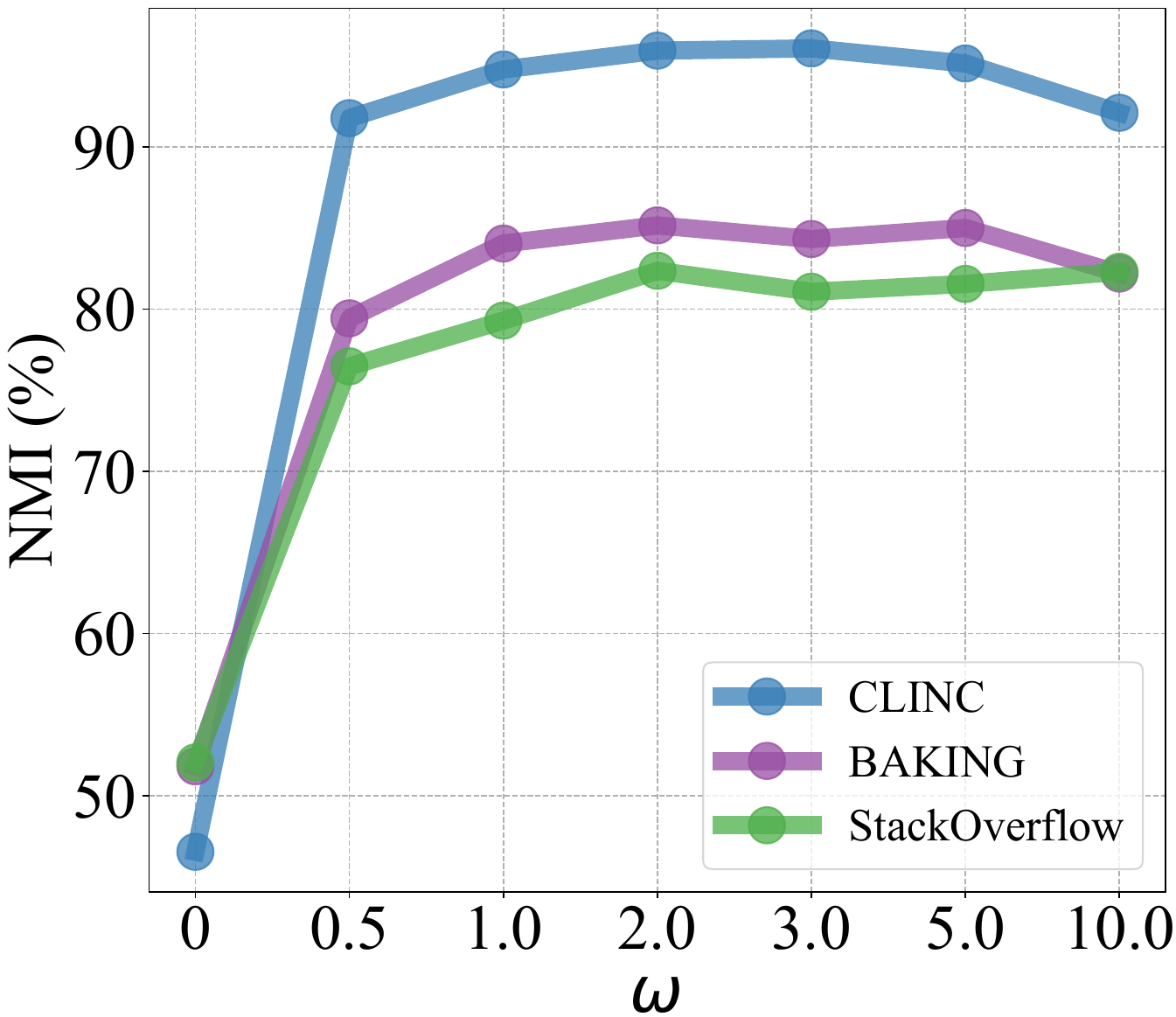}
    \label{subfigure: scatter_dac2}
    }
    \caption{Weight of the multitask learning $\omega$.} 
    \vspace{-13pt}
    \label{fig:fig_w}
\end{figure}

\subsection{Weight of Multitask Learning}
The weight of the multitask learning $\omega$ in Eq.~\ref{eq:11} adjusts the contribution of two objectives RPAL and APDL.
To pursue the optimal performance, we conduct experiments varying $\omega$ across $\{0.0, 0.5, 1.0, 2.0, 3.0, 5.0, 10.0\}$. 
In Figure~\ref{fig:fig_w}, the model performance continues improving with the value of $\omega$ increasing to $1.0$. The model keeps a relatively stable performance after reaching $3.0$ but encounters performance degradation with a large value of $\omega=10.0$. The ability of the RPAL cannot be fully exploited with a small value of $\omega$, while the ability of the APDL is suppressed leading to a worse clustering performance with a large value of $\omega$.
Empirically, we choose $\omega=2.0$ for all datasets.

\subsection{Comparison with LLM}
To conduct a comprehensive performance comparison with the large language model (LLM) on the NID task, we randomly select $1.5\%$ of training data as labeled and choose $75\%$ of all intents as known. For evaluation, our comprehensive assessment covers 600 instances from three datasets (200 samples randomly from each dataset).
The metrics are comprised of known intents accuracy (KACC), novel detection accuracy (NACC), and clustering performance (NMI and ARI) for novel intents. 
In Table~\ref{tab:with_LLM}, our method consistently outperforms ChatGPT3.5\footnote{We utilize OpenAI \emph{gpt-3.5-turbo-0301}, see \href{https://platform.openai.com/docs/models/overview}{index}.} across all datasets and evaluation metrics with a small model size and fast inference speed, demonstrating the superior performance of our approach. Moving forward, we plan to explore the integration of \ourmethod{} with LLM to boost the performance in NID.

\begin{table}[t]
    \centering
    \resizebox{0.95\columnwidth}{!}{
    \begin{tabular}{l | c | c | c | c}
    \toprule
     \textbf{Methods} & \textbf{KACC} & \textbf{NACC} & \textbf{NMI} & \textbf{ARI}\\
    \midrule
    ChatGPT3.5 & \textbf{\underline{84.14}} & 25.45 & 73.58 & 37.88 \\ 
    \textit{Ours} \texttt{(CLINC)} & 68.56 & \textbf{\underline{82.00}}& \textbf{\underline{88.53}} & \textbf{\underline{47.06}}\\
    \arrayrulecolor{lightgray}
    \midrule
    \midrule
    \arrayrulecolor{black}
    ChatGPT3.5 & \textbf{\underline{80.41}} & 32.69 & 87.69 & 46.68 \\ 
    \textit{Ours} \texttt{(BANKING)} & 71.42 & \textbf{\underline{85.00}}& \textbf{\underline{89.60}} & \textbf{\underline{58.66}}\\
    \arrayrulecolor{lightgray}
    \midrule
    \midrule
    \arrayrulecolor{black}
    ChatGPT3.5 & 71.53 & 50.00 & 72.20 & 47.17 \\ 
    \textit{Ours} \texttt{(StackOverflow)} & \textbf{\underline{82.35}} & \textbf{\underline{89.41}}& \textbf{\underline{81.50}} & \textbf{\underline{62.13}}\\
    \bottomrule
    \end{tabular}
    }
    \caption{Comparison between \ourmethod{} and LLM.}
    \vspace{-10pt}
    \label{tab:with_LLM}
\end{table}

\section{Conclusion}
\label{sec:Conclusion}
In this work, we propose a robust and adaptive prototypical learning (\ourmethod{}) framework for new intent discovery, which aims to learn cluster-friendly discriminative representations.
Specifically, we design the robust prototypical attracting learning (RPAL) method and the adaptive prototypical dispersion (APDL) method to control within-cluster and between-cluster distances, respectively.
Experimental results on three benchmarks demonstrate that \ourmethod{} significantly outperforms the previous unsupervised and semi-supervised baselines and even defeats the large language model (ChatGPT3.5). 
Extensive probing analysis further verifies that RPAL is helpful for realizing stronger within-cluster compactness while mitigating the effects of noisy pseudo-labels and APDL is beneficial for attaining larger between-cluster dispersion.
We hope our work can provide useful insights for further research.

\clearpage
\section*{Limitations}
Despite the promising results obtained by our~\ourmethod{}, it is crucial to acknowledge several limitations:
\textbf{(1) Improving pseudo-labels assignments.} The pseudo-labels obtained using the k-means method are not sufficiently reliable, as it is highly sensitive to noisy intent data.
We plan to explore more reliable pseudo-label assignment approaches.
\textbf{(2) Leveraging LLMs to facilitate interpretability.} While our clustering method can assign cluster labels to unlabeled utterances, it cannot generate meaningful and interpretable names for each identified cluster or intent. 
We intend to investigate the combination of our method with LLMs to assign accurate category names to newly discovered intent categories.
%

\section*{Acknowledgements} This work was supported in part by the National Natural Science Foundation of China (Grant Nos. U1636211, U2333205, 61672081, 62302025, 62276017), a fund project: State Grid Co., Ltd. Technology R\&D Project (ProjectName: Research on Key Technologies of Data Scenario-based Security Governance and Emergency Blocking in Power Monitoring System, Proiect No.: 5108-202303439A-3-2-ZN), the 2022 CCF-NSFOCUS Kun-Peng Scientific Research Fund and the Opening Project of Shanghai Trusted Industrial Control Platform and the State Key Laboratory of Complex \& Critical Software Environment (Grant No. SKLSDE-2021ZX-18).


\nocite{*}
\section{Bibliographical References}\label{sec:reference}

\bibliographystyle{lrec-coling2024-natbib}
\bibliography{lrec-coling2024-example}


\end{document}